\documentclass[10pt,twocolumn,letterpaper]{article}

\usepackage{cvpr}
\usepackage{epsfig}
\usepackage{graphicx}
\usepackage{amsmath}
\usepackage{amssymb}
\usepackage{xcolor}
\RequirePackage{fontawesome}

\usepackage{color, colortbl}
\definecolor{Gray}{gray}{0.94}

\usepackage[pagebackref,breaklinks,colorlinks]{hyperref}

\iftrue
\newcommand{\HC}[1]{\textcolor{green}{\textbf{HC}:~#1}}
\newcommand{\CL}[1]{\textcolor{blue}{\textbf{CL}:~#1}}
\newcommand{\TS}[1]{{\textcolor{green}{\textbf{TS}:~#1}}}
\newcommand{\ET}[1]{\textcolor{blue}{\textbf{ET}:~#1}}
\newcommand{\LK}[1]{\textcolor{orange}{\textbf{LK}:~#1}}
\newcommand{\EV}[1]{\textcolor{orange}{\textbf{EV}:~#1}}
\newcommand{\PK}[1]{\textcolor{cyan}{\textbf{PK}:~#1}}
\else
\newcommand{\HC}[1]{}
\newcommand{\CL}[1]{}
\newcommand{\TS}[1]{}
\newcommand{\ET}[1]{}
\newcommand{\LK}[1]{}
\newcommand{\EV}[1]{}
\newcommand{\PK}[1]{}
\fi

\newcommand{\methodname}[0]{VEO}
\newcommand{\lame}[0]{Lam\'{e} }
\DeclareMathOperator*{\argmin}{arg\,min}

\def\cvprPaperID{}
\def\confName{}
\def\confYear{}

\begin{document}

\title{Virtual Elastic Objects}

\author{\normalsize Hsiao-yu Chen$^{1,3*}$, Edgar Tretschk$^{2,3*}$, Tuur Stuyck$^3$, Petr Kadlecek$^3$, Ladislav Kavan$^3$, Etienne Vouga$^1$, Christoph Lassner$^3$\\
\normalsize$^1$University of Texas at Austin, $^2$Max Planck Institute for Informatics, $^3$Meta Reality Labs Research\\
}

\twocolumn[{%
\renewcommand\twocolumn[1][]{#1}%
\maketitle
\vspace{-0.5cm}
\centering
    \hspace*{0.5cm}\includegraphics[width=0.9\textwidth]{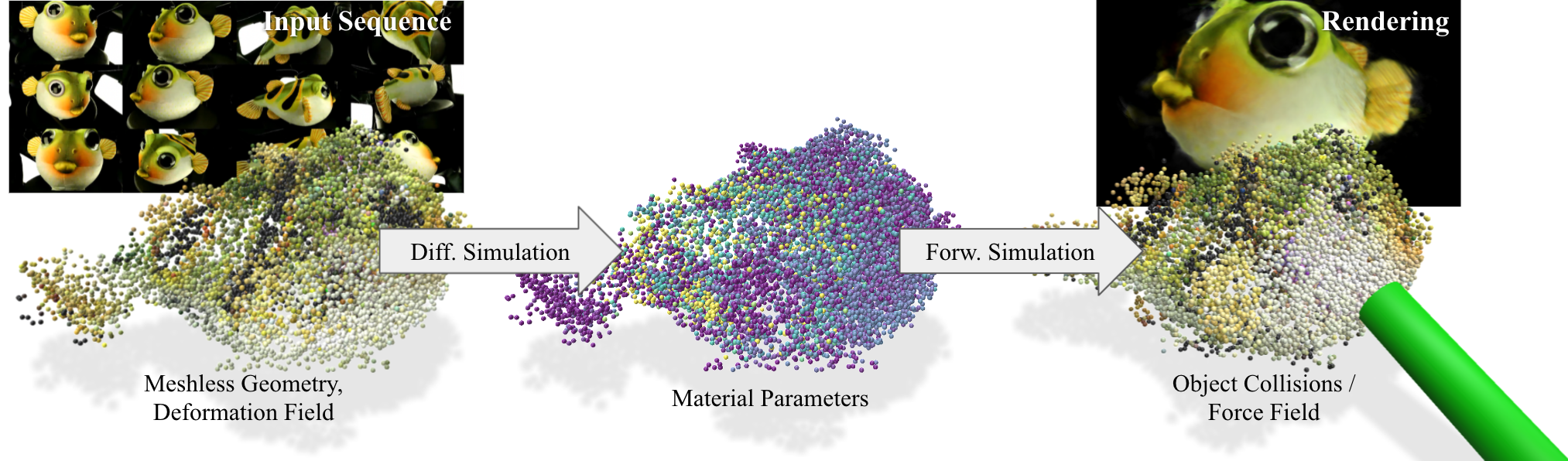}
    \vspace*{-0.3cm}
    \captionof{figure}{
        \textit{Method overview.} We use a multi-view capture system to record objects deforming under the influence of external forces.
        Our method reconstructs a meshless geometry and deformation field from these sequences.
        Using a differentiable simulator, we optimize the material parameters of the objects to match the observations.
        These parameters allow us to find novel, plausible object configurations in response to new forces fields or collision constraints due to user interactions
        Finally, we re-render the deformed state.
    }
    \vspace*{0.2cm}
    \label{fig:teaser}
}]

\begin{abstract}
We present \emph{Virtual Elastic Objects} (VEOs): virtual objects that not only look like their real-world counterparts but also behave like them, even when subject to novel interactions. 
Achieving this presents multiple challenges: not only do objects have to be captured including the physical forces acting on them, then faithfully reconstructed and rendered, but also plausible material parameters found and simulated.
To create VEOs, we built a multi-view capture system that captures objects under the influence of a compressed air stream.
Building on recent advances in model-free, dynamic Neural Radiance Fields, we reconstruct the objects and corresponding deformation fields.
We propose to use a differentiable, particle-based simulator to use these deformation fields to find representative material parameters, which enable us to run new simulations.
To render simulated objects, we devise a method for integrating the simulation results with Neural Radiance Fields.
The resulting method is applicable to a wide range of scenarios: it can handle objects composed of inhomogeneous material, with very different shapes, and it can simulate interactions with other virtual objects.
We present our results using a newly collected dataset of 12 objects under a variety of force fields, which will be shared with the community.

\vspace*{0.2cm}
\noindent\emph{$^*$Work was done while Hsiao-yu (partially) and Edgar were interning at Reality Labs Research.}

\end{abstract}

\section{Introduction}

3D reconstruction is one of the fundamental problems of computer vision and a cornerstone of augmented and virtual reality. 
Concurrently with steady progress towards real-time photo-realistic rendering of 3D environments in game engines,
the last few decades have seen great strides towards photo-realistic 3D reconstruction. 
A recent achievement in this direction is the discovery of a fairly general formulation for representing radiance fields \cite{Mildenhall_2020_NeRF,liu2020neural,martin2021nerf,Schwarz2020NEURIPS,zhang2020nerf++,yu2021pixelnerf,trevithick2020grf,bi2020neural,srinivasan2021nerv,niemeyer2021giraffe,sucar2021imap}. 
Neural radiance fields are remarkably versatile for reconstructing real-world objects with high-fidelity \emph{geometry} and \emph{appearance}. 
But static appearance is only the first step: it ignores how an object moves and interacts with its environment. 
4D reconstruction tackles this problem in part by incorporating the time dimension: with more intricate capture setups and more data, we can reconstruct objects over time---but can only re-play 
the captured sequences. 
Today, in the age of mixed reality, a photo-realistically reconstructed object might still destroy immersion if it is not ``physically realistic'' because \emph{the object cannot be interacted with.} 
(For example, if a soft object appears as rigid as the rocks next to it when stepped on.)

By building on advances in computer vision and physics simulation, we begin to tackle the problem of physically-realistic reconstruction and create \emph{Virtual Elastic Objects}: virtual objects that not only look like their real-world counterparts but also behave like them, even when subject to novel interactions. For the first time, this allows for full-loop reconstruction of deforming elastic objects: from capture, to reconstruction, to simulation, to interaction, to re-rendering.

Our core observation is that with the latest advances in 4D reconstruction using neural radiance fields, we can both capture radiance and deformation fields of a moving object over time, and re-render the object given novel deformation fields. That leaves as the main challenge the core problem of capturing an object's physics from observations of its interactions with the environment. With the right representation that jointly encodes an object's geometry, deformation, and material behavior, compatible with both differentiable physical simulation and the deformation fields provided by 4D reconstruction algorithms, we can use these deformation fields to provide the necessary supervision to learn the material parameters.

But even with this insight, multiple challenges remain to create Virtual Elastic Objects. We list them together with our technical contributions:\\
\noindent\textbf{1) Capture.} 
To create VEOs, we need to collect data that not only contains visual information but also information about physical forces. 
We present the new \textbf{PLUSH} dataset containing occlusion-free 4D recordings of elastic objects deforming under known controlled force fields. 
To create this dataset, we built a multi-camera capture rig that incorporates an air compressor with a movable, tracked nozzle. 
More details can be found in Sec.~\ref{ssec:capture}. \\
\noindent\textbf{2) Reconstruction.} 
VEOs~do not require any prior knowledge about the geometry of the object to be reconstructed; the reconstruction thus must be template-free and provide full 4D information (\ie, a 3D reconstruction and deformation information over time). 
We extend Non-rigid Neural Radiance Fields~\cite{tretschk2021nonrigid} with novel losses, and export point clouds and point correspondences to create the data required to supervise learning material behavior using physical simulation.
We provide further details in Sec.~\ref{ssec:recon}.\\
\noindent\textbf{3) Simulation.} 
Crucially for creating realistic interactive objects, a physical simulation is required, both to optimize for an unknown object's physical parameters and to generate deformations of that object in response to novel interactions. 
We implement a differentiable quasi-static simulator that is particle-based and is compatible with the deformation field data provided by our 4D reconstruction algorithm.
We present the differentiable simulator and explain how we use it to obtain physical parameters in Sec.~\ref{ssec:learning_material}, and describe simulations of novel interactions in Sec.~\ref{ssec:interaction}.\\ 
\noindent\textbf{4) Rendering.} 
Since we convert from a neural representation of the captured object's geometry to a point cloud reconstructing the object's physical properties, we require a function that allows rendering the object given new simulated deformations of the point cloud. 
We introduce a mapping function that enables us to use deformed point clouds instead of continuous deformation fields to alter the ray casting for the Neural Radiance Fields we used for the original reconstruction. 
Further details on re-rendering can be found in Sec.~\ref{ssec:rendering}.

\section{Related Work}
Our work integrates together multiple areas of computer vision, computer graphics, and simulation. 

\noindent\textbf{Recovering Elastic Parameters for 3D Templates.}
A number of prior works estimate material parameters of a pre-scanned 3D template by tracking the object over time from depth input. 
Wang~\etal~\cite{wang2015deformation} were among the first to tackle tracking, rest pose estimation, and material parameter estimation from multi-view depth streams. They adopt a gradient-free downhill simplex method for parameter fitting, and can only optimize a limited number of material parameters. Objects built from multiple types of materials cannot be faithfully captured without manual guidance or prior knowledge of a part decomposition. 
Hahn~\etal~\cite{Hahn:2019} learn an inhomogeneous viscoelastic model from recordings of motion markers covering the object.
Recently, Weiss~\etal~\cite{weiss2020correspondence} infer homogeneous linear material properties by tracking deformations of a given template with a single depth camera. 
In contrast to these methods, ours jointly reconstructs not just object deformations and physics \emph{without a need for depth input or markers} but also geometry and appearance \emph{without a need for a template}. Our formulation can model inhomogeneous, nonlinear materials without prior knowledge or annotations. 

\noindent\textbf{Dynamic Reconstruction.}
Reconstructing non-rigid objects from a video sequence is a long-standing computer vision and graphics problem~\cite{Zhang2003SpacetimeSS, tung_complete_2009}. 
Shape-from-Template methods deform a provided template using RGB~\cite{yu2015direct} or RGB-D data~\cite{zollhofer2014real}. 
DynamicFusion~\cite{Newcombe_2015_CVPR} is a model-free, real-time method for reconstructing general scenes from a single RGB-D video.
When reliable 2D correspondences are available from optical flow, non-rigid structure-from-motion (NRSfM) can be used to reconstruct the 3D geometry~\cite{Agudo2014OnlineDN, grassmanian_2018}, perhaps even using physics-based priors~\cite{agudo2015sequential}. 
There are also image-based approaches that do not yield a true 3D scene~\cite{yoon2020novel,Bemana2020xfields}. 
Recently, reconstruction using neural representations have become more common. 
Whereas OccupancyFlow~\cite{niemeyer2019occupancy} requires 3D supervision,
Neural Volumes~\cite{Lombardi:2019} reconstructs a dynamic scene from multi-view input only, but does not compute temporal correspondences. See a recent survey on neural rendering~\cite{Tewari2020NeuralSTAR} for more. 

Neural Radiance Fields~\cite{Mildenhall_2020_NeRF}, the seminal work of Mildenhall~\etal, lays the groundwork for several follow-up reconstruction methods that extend it to dynamic scenes~\cite{Li2021, park2021hypernerf,attal2021torf,pumarola2020d,park2021nerfies,li2021neural,du2021nerflow,Gaofreeviewvideo,xian2021space,Lombardi_2021_MVP}. In this work, we assume multi-view RGB video input with known camera parameters and foreground segmentation masks and so extend Non-Rigid Neural Radiance Fields (NR-NeRF)~\cite{tretschk2021nonrigid}.

\noindent\textbf{Data-Driven Physics Simulation.}
Much recent research has explored the potential of machine learning to enhance or even replace traditional physics-based simulation. Learning natural laws from data without any priors has been shown possible for a few simple physics systems\cite{schmidt2009distilling}, but the computational cost scales exponentially with the complexity of the system, and remains intractable for real-world problems. For simulating elastic objects specifically, one line of work replaces traditional mesh kinematics with a learned deformation representation to improve performance: Fulton~\etal~\cite{Fulton:2018} use an autoencoder to learn a nonlinear subspace for elastic deformation, and Holden~\etal~\cite{Holden:2019} train a neural network to predict the deformation of cloth using a neural subspace. Some methods use neural networks to augment coarse traditional simulations with fine details~\cite{deepWrinkles,geng2020coercing}. 

Another line of work uses data to fit a parameterized material model to observed deformations. This idea has been successfully applied to muscle-actuated biomechanical systems such as human faces~\cite{Kadlecek:2019,Srinivasan:2021}, learning the rest pose of an object in zero gravity~\cite{chen2014anm}, the design of soft robotics~\cite{hu2019chainqueen, hu2020difftaichi}, and motion planning with frictional contacts~\cite{Geilinger:2020,du2021diffpd}. Yang~\etal~\cite{Yang2017} learn physical parameters for cloth by analysing the wrinkle patterns in video. While all of these methods learn physical parameters from data, our method is unique in requiring no template or other prior knowledge about object geometry to reconstruct and re-render novel deformations of an object.

\noindent\textbf{Meshless Simulation.}
Meshless physics-based simulation emerged as a counter-part to traditional mesh-based methods \cite{muller2004} and is ideal for effects such as melting or fracture \cite{muller2004,pauly2004meshless}. 
These methods have been later extended to support oriented particles and skinning \cite{muller2011solid,gilles2011frame,macklin2014unified}. 
Another extension of point-based simulations consists in incorporating a background Eulerian grid, which enables more efficient simulation of fluid-like phenomena \cite{stomakhin2013material,jiang2017anisotropic}.
\section{Method}

\subsection{Capture}
\label{ssec:capture}

To create a physically accurate representation of an object, we first need to record visual data of its deformation under known physical forces. 
For recording, we use a static multi-view camera setup consisting of 19 OpenCV AI-Kit Depth (OAK-D) cameras\footnote{\url{https://store.opencv.ai/products/oak-d}}, each containing an RGB and two grey-scale cameras (note that \methodname~does not use the stereo camera data to infer classical pairwise stereo depth). 
They represent an affordable, yet surprisingly powerful solution for volumetric capture. 
In particular, their on-board H265 encoding capability facilitates handling the amount of data produced during recording (5.12GB/s uncompressed). 
Since the cameras lack a lens system with zoom capabilities, we keep them close to the object to optimize the pixel coverage and re-configure the system depending on object size. 
The maximum capture volume has a size of roughly $30\text{cm}^3$. 
We put a black sheet around it to create a dark background with the exception of five stage lights that create a uniform lighting environment.

In addition to the images, we also need to record force fields on the object surface. 
This raises a problem: if a prop is used to exert force on the capture subject, the prop becomes an occluder that interferes with photometric reconstruction. We solved this problem when capturing our \textbf{PLUSH} dataset by actuating the object using transparent fishing line and a compressed air stream; see Sec.~\ref{sec:dataset} for further details.

\subsection{4D Reconstruction}
\label{ssec:recon}

Given the captured video of an object deforming under external forces, we need 4D reconstruction to supply a temporally-coherent point cloud that can be used to learn the object material properties.
To that end, we use NR-NeRF~\cite{tretschk2021nonrigid}, which extends the static reconstruction method NeRF~\cite{Mildenhall_2020_NeRF} to the temporal domain. 
NeRF learns a volumetric scene representation: a coordinate-based Multi-Layer Perceptron (MLP) $\mathbf{v}(\mathbf{x}) = (o, \mathbf{c})$ that regresses geometry (opacity $o(\mathbf{x})\in\mathbb{R}$) and appearance (RGB color $\mathbf{c}(\mathbf{x})\in\mathbb{R}^3$) at each point $\mathbf{x}$ in 3D space. 
At training time, the weights of $\mathbf{v}$ are optimized through 2D supervision by RGB images with known camera parameters: for a given pixel of an input image, the camera parameters allow us to trace the corresponding ray $\mathbf{r}(s)$ through 3D space. 
We then sample the NeRF at $|S|$ points $\{\mathbf{r}(s)\in\mathbb{R}^3\}_{s \in S}$ along the ray, and use a volumetric rendering equation to accumulate the samples front-to-back via weighted averaging: $\Tilde{\mathbf{c}} = \sum_{s\in S} \alpha_s \mathbf{c}(\mathbf{r}(s))$ (\ie, alpha blending with alpha values $\{\alpha_s\in\mathbb{R}\}_s$ derived from the opacities $\{o_s\}_s$). 
A reconstruction loss encourages the resulting RGB value $\Tilde{\mathbf{c}}$ to be similar to the RGB value of the input pixel. 

On top of the static geometry and appearance representation $\mathbf{v}$ (the \emph{canonical model}), NR-NeRF models deformations explicitly via a jointly learned ray-bending MLP $\mathbf{b}(\mathbf{x},\mathbf{l}_t) = \mathbf{d} $ that regresses a 3D offset $\mathbf{d}$ for each point in space at time $t$. 
($\mathbf{l}_t$ is an auto-decoded latent code that conditions $\mathbf{b}$ on the deformation at time $t$.) 
When rendering a pixel at time $t$ with NR-NeRF, $\mathbf{b}$ is queried for each sample $\mathbf{r}(s)$ on the ray in order to deform it into the canonical model: $(o,\mathbf{c}) = \mathbf{v}\left[\mathbf{r}(s) + \mathbf{b}(\mathbf{r}(s),\mathbf{l}_t)\right]$. 
Unlike NR-NeRF's monocular setting, we have a multi-view capture setup. 
We thus disable the regularization losses of NR-NeRF and only use its reconstruction loss. 

\noindent\textbf{Extensions.} 
We improve NR-NeRF in several ways to adapt it to our setting.
The input videos contain background, which we do not want to reconstruct. 
We obtain foreground segmentations for all input images via image matting~\cite{Lin2021} together with a hard brightness threshold. 
During training, we use a background loss $L_\mathit{background}$ to discourage geometry along rays of background pixels. 
When later extracting point clouds, we need opaque samples on the inside of the object as well. 
However, we find that $L_\mathit{background}$ leads the canonical model to prefer empty space even inside the object.
We counteract this effect with a density loss $L_\mathit{density}$ that raises the opacity of point samples of a foreground ray that are `behind' the surface, while emptying out the space in front of the surface with $L_\mathit{foreground}$. 
During training, we first build a canonical representation by pretraining the canonical model on a few frames and subsequently using it to reconstruct all images. 
Our capture setup not only provides RGB streams but also grey-scale images. 
We use these for supervision as well. 
In practice, we use a custom weighted combination of these techniques for each sequence to get the best reconstruction. 

\noindent\textbf{Point Cloud Extraction} 
In order to extract a temporally-consistent point cloud from this reconstruction, we require a forward deformation model, which warps from the canonical model to the deformed state at time~$t$. 
However, NR-NeRF's deformation model~$\mathbf{b}$ is a backward warping model: it deforms each 
deformed state into the canonical model. 
We therefore jointly train a coordinate-based MLP $\mathbf{w}$ to approximate the inverse of $\mathbf{b}$.
After training, we need to convert the reconstruction from its continuous MLP format into an explicit point cloud. 
To achieve that, we cast rays from all input cameras and extract points from the canonical model that are at or behind the surface and whose opacity exceeds a threshold. 
These points can then be deformed from the canonical model into the deformed state at time $t$ via $\mathbf{w}$.
We thus obtain a 4D reconstruction in the form of a 3D point cloud's evolving point positions $\{P_t\}_t$, which are in correspondence across time. 
To keep the computational cost of the subsequent reconstruction steps feasible, we downsample the point cloud to 9-15$k$ points if necessary.

\subsection{Learning Material Parameters}
\label{ssec:learning_material} 
Before we can simulate novel interactions with a captured object, we need to infer its physical behavior. Given that we have no prior knowledge of the object, we make several simplifying assumptions about its mechanics, with an eye towards minimizing the complexity of the physical model while also remaining flexible enough to capture heterogeneous objects built from multiple materials. 

First, we assume a \emph{spatially varying, isotropic nonlinear Neo-Hookean material model} for the object. Neo-Hookean elasticity well-approximates the behavior of many real-world materials, including rubber and many types of plastic, and is popular in computer graphics applications because its nonlinear stress-strain relationship guarantees that no part of the object can invert to have negative volume, even if the object is subjected to arbitrary large and nonlinear deformations.
Finally, Neo-Hookean elasticity admits a simple parameterization: a pair of \lame parameters  $(\mu_i, \lambda_i)\in\mathbb{R}^2$ at each point $i$ of the point cloud $P$.

Second, we assume that the object deforms \emph{quasistatically} over time: that at each point in time, the internal elastic forces exactly balance gravity and applied external forces. The quasistatic assumption greatly simplifies learning material parameters, and is valid so long as inertial forces in the captured video sequences are negligible (or equivalently, so long as external forces change sufficiently slowly over time that there is no secondary motion, which is true for the air stream and string actuations in our \textbf{PLUSH} dataset).

\paragraph{Overview.} We first formulate a differentiable, mesh-free \emph{forward} physical simulator that is tailored to work directly with the (potentially noisy) reconstructed point cloud. This forward simulator maps from the point cloud $P_0$ of the object in its \emph{reference pose} (where it is subject to no external forces besides gravity), an assignment of \lame parameters to every point, and an assignment of an external force $\mathbf{f}_i\in\mathbb{R}^3$ to each point on the object surface, to the deformed position $\mathbf{y}_i\in\mathbb{R}^3$ of every point in the point cloud after the object equilibrates against the applied forces.

Next, we learn the \lame parameters that match the object's observed behavior by minimizing a loss function $\mathbf{L}$ that sums, over all times $t$, the distance between $\mathbf{y}_i$ and the corresponding target position of the point in the 4D point cloud $P_t$.

\paragraph{Quasistatic Simulation.} 
To compute the equilibrium positions $\mathbf{y}_i$ of the points in $P$ for given external loads and material parameters, we solve the variational problem
\begin{equation}
    \argmin_\mathbf{y} \mathbf{E}(\mathbf{y}), \label{eq:equilibrium}
\end{equation}
where $\mathbf{E}$ is the total energy of the physical system, capturing both the elastic energy of deformation as well as work done on the system by external forces. In what follows, we derive the expression for $\mathbf{E}$, and discuss how to solve Eq.~\ref{eq:equilibrium}.

Following M\"{u}ller \etal~\cite{muller2004}, we adopt a mesh-free, point-based discretization of elasticity to perform forward simulation. For every point $\mathbf{x}_i$ in the reference point cloud $P_0$, we define a neighborhood $\mathcal{N}_i$ containing the 6 nearest neighbors of $\mathbf{x}_i$ in $P_0$. For any given set of deformed positions $\mathbf{y}_j$ of the points in $\mathcal{N}_i$, we estimate strain within the neighborhood in the least-squares sense. More specifically, the local material deformation gradient $\mathbf{F}_i\in\mathbb{R}^3$ maps the neighborhood $\mathcal{N}_i$ from the reference to the deformed state:
\begin{equation}
    \label{eq:deformation_gradient}
    \mathbf{F}_i(\mathbf{x}_i-\mathbf{x}_j) \approx \mathbf{y}_i-\mathbf{y}_j \quad \forall \mathbf{x}_j \in \mathcal{N}_i.
\end{equation}
For neighborhoods larger than three, Eq.~\ref{eq:deformation_gradient} is over-determined, and we hence solve for $\mathbf{F}_i$ in the least-squares sense,
yielding the closed-form solution:
\begin{equation}
    \mathbf{F}_i = \mathbf{Y}_i \mathbf{W}_i \mathbf{X}_i^T (\mathbf{X}_i \mathbf{W}_i \mathbf{X}_i^T)^{-1},
\end{equation}
where the $j$-th column of $\mathbf{X}_i$ and $\mathbf{Y}_i$ are $\mathbf{x}_i - \mathbf{x}_j$ and $\mathbf{y}_i - \mathbf{y}_j$, respectively, and $\mathbf{W}_i$ is a diagonal matrix of weights depending on the distance from $\mathbf{x}_j$ to $\mathbf{x}_i$~\cite{muller2004}. 

The elastic energy of the object can be computed from the classic Neo-Hookean energy density~\cite{ogden1984non}:
\begin{equation}
    \label{eq:neoHookean}
    \Psi_\mathit{NH}^i = \frac{\mu_i}{2}(I_c -3 ) - \mu_i \log J + \frac{\lambda_i}{2}(J-1)^2,
\end{equation}
where $I_c$ is the trace of the right Cauchy-Green tensor $\mathbf{F}_i^T\mathbf{F}_i$, and $J$ is the determinant of $\mathbf{F}_i$.
$\mu_i$ and $\lambda_i$ are the \lame parameters assigned to point $i$.
The total elastic energy is then:
\begin{equation}
    \mathbf{E}_\mathit{NH} = \sum_i V_i \Psi_\mathit{NH}^i,
\end{equation}
where $V_i\in\mathbb{R}$ approximates the volume of $\mathcal{N}_i$. 

We also need to include the virtual work done by the external force field to Eq.~\ref{eq:equilibrium}:
\begin{equation}
    \mathbf{E}_W = \sum_i \mathbf{f}_i \cdot \mathbf{y}_i, \label{eq:virtualwork}
\end{equation}
where $\mathbf{f_i}$ is the force applied to point $i$ (the force of the air stream on the boundary). If we measured the tension in the fishing lines, we could also include the forces they exert on the object in Eq.~\ref{eq:virtualwork}. But since a fishing line is effectively inextensible relative to the object we are reconstructing, we instead incorporate the fishing lines as soft constraints on the positions of the points $Q\subset P$ attached to the lines: we assume that at time $t$, points in $Q$ should match their observed positions in $P_t$, and formulate an attraction energy:
\begin{equation}
    \mathbf{E}_A = \alpha \sum_{q \in Q} \lVert \mathbf{y}_q - \mathbf{x}_q^* \rVert^2,
\end{equation}
where $\mathbf{x}_q^*$ is the position of the point corresponding to $\mathbf{y}_q$ in $P_t$, and $\alpha$ is a large penalty constant. We found that this soft constraint formulation works better in practice than alternatives such as enforcing $\mathbf{y}_q = \mathbf{x}_q^*$ as a hard constraint.

The total energy in Eq.~~\ref{eq:equilibrium} is thus $\mathbf{E} = \mathbf{E}_{NH} + \mathbf{E}_W + \mathbf{E}_A$, which we minimize using Newton's method. Since Newton's method can fail when the Hessian $\mathbf{H}$ of $\mathbf{E}$ is not positive-definite, we perform a per-neighborhood eigen-decomposition of $\mathbf{H}$ and replace all eigenvalues that are smaller than a threshold $\epsilon>0$ with $\epsilon$; note that this is a well-known technique to improve robustness of physical simulations~\cite{teran2005robust}. We also make use of a line search to ensure stability and handling of position constraints at points where the capture subject touches the ground; see the supplemental material for further implementation details.

\paragraph{Material Reconstruction.} 
Given the 4D point cloud $P_t$ and forces acting on the object $\{\mathbf{f}_i\}_i$, we use our forward simulator to learn the \lame parameters that best explain the observed deformations. More specifically, at each time $t$ we define the loss:
\begin{equation}
    \mathbf{L}_t = \sum_{i \in \partial \Omega} \lVert \mathbf{y}_{t,i} - \mathbf{x}_{t,i}^*
    \rVert^2 
\end{equation}
where $\mathbf{x}_{t,i}^*$ is the position of point $i$ in $P_t$, and $\mathbf{y}_{t,i}$ is the output of the forward simulation. We use an $\ell_2$ loss to penalize outliers strongly, which would jeopardize the reconstruction quality otherwise. 

We choose a training subsequence $T$ of 20-50 frames from the input where the impact of the air stream roughly covers the surface so that we have some reference for each part of the object, and compute the desired \lame parameters by minimizing the sum of the loss over all $t \in T$ using the gradient-based Adam optimizer~\cite{KingmaB14}:
\begin{equation}
    \mu^*, \lambda^* = \argmin_{\mu, \lambda} \sum_{t\in T} \mathbf{L}_t.
\end{equation}

It is not trivial to back-propagate through the Newton solve for $\mathbf{y}_{t,i}$, even if we ignore the line search and assume a fixed number of Newton iterations $K$. The gradient of $\mathbf{y}$ with respect to the \lame parameters ($\mu$ for instance) can be computed using the chain rule:
\begin{equation}
    \label{eq:devLoss}
    \frac{\partial \mathbf{L}}{\partial \mu} = \frac{\partial \mathbf{L}}{\partial \mathbf{y}^K}\frac{\partial \mathbf{y}^K}{\partial \mu},
\end{equation}
and, for any $1 \leq k \leq K$,
\begin{align}
        \frac{\partial \mathbf{y}^k}{\partial \mu} &= \frac{\partial \mathbf{y}_{k-1}}{\partial \mu} -\left(\frac{\partial\mathbf{H}_{k-1}^{-1}}{\partial \mu} + \frac{\partial\mathbf{H}_{k-1}^{-1}}{\partial \mathbf{y}_{k-1}}\frac{\partial \mathbf{y}_{k-1}}{\partial \mu}\right)\nabla\mathbf{E}_{k-1}   \notag\\   & -\mathbf{H}_{k-1}^{-1}\left(\frac{\partial\nabla\mathbf{E}_{k-1}}{\partial \mu} + \frac{\partial\nabla\mathbf{E}_{k-1}}{\partial \mathbf{y}_{k-1}}\frac{\partial \mathbf{y}_{k-1}}{\partial \mu}\right).
\end{align}
To avoid an exponentially-large expression tree, we approximate the derivative of the $k$th Newton iterate $\mathbf{y}^k$ by neglecting the higher-order derivative of the Hessian and of the gradient of the energy with respect to the previous position update:
\begin{align*}
         \frac{\partial \mathbf{y}^k}{\partial \mu} \approx \frac{\partial \mathbf{y}_{k-1}}{\partial \mu} -\frac{\partial\mathbf{H}_{k-1}^{-1}}{\partial \mu} \nabla\mathbf{E}_{k-1} -\mathbf{H}_{k-1}^{-1} \frac{\partial\nabla\mathbf{E}_{k-1}}{\partial \mu}
\end{align*}
Although it is not guaranteed that the higher-order terms are always negligible, this approximation provides a sufficiently high-quality descent direction for all examples we tested. To improve performance and to capture hysteresis in cases where $\mathbf{E}$ has multiple local minima at some times $t$, we warm-start the Newton optimization at time $t$ using the solution from time $t-1$.

\subsection{Novel Interactions}
\label{ssec:interaction}
Given a reconstructed \methodname, we can use the same physical simulator used for material inference to re-simulate the captured object subject to novel interactions. New force fields can easily be introduced by modifying $\mathbf{f}_i$ in the energy $\mathbf{E}_W.$ Other possible interactions include changing the direction of gravity, adding contact forces to allow multiple objects to mutually interact, or to allow manipulation of the object using mixed-reality tools, etc.

We demonstrate the feasibility of re-simulating novel interactions by implementing a simple penalty energy to handle contact between a \methodname~and a secondary object, represented implicitly as a signed distance field $d:\mathbb{R}^3\to\mathbb{R}$. The penalty energy is given by:
\begin{align}
  \Psi_c(\mathbf{y}) &=
    \begin{cases}
     \alpha_c d(\mathbf{y})^2 & \text{if } d(\mathbf{y}) < 0\\
      0 & \text{otherwise,}\\
    \end{cases} \\
    \mathbf{E}_c &= \sum_i V_i \Psi_c(\mathbf{y}_i),
\end{align}
where $\alpha_c$ is chosen large enough to prevent visually-noticeable penetration of the VEO by the secondary object.

\subsection{Rendering} 
\label{ssec:rendering}
We are able to interact freely with the \methodname~in a physically plausible manner. 
Hence, we can close the full loop and realistically render the results of simulated novel interactions using neural radiance fields. 
While we used $\mathbf{b}$ for deformations during the reconstruction, we are now given a new deformed state induced by a discrete point cloud: a canonical reference point cloud $P_0 = \{ \mathbf{x}^0_{s} \}_s$ and its deformed version $S_d = \{ \mathbf{y}^d_{s} \}_s$. 
We need to obtain a continuous backward-warping field from that point cloud in order to replace $\mathbf{b}$, which bends straight rays into the canonical model. 
To that end, we interpolate the deformation offsets $\mathbf{d}^b_s = \mathbf{x}^0_{s} - \mathbf{y}^d_{s}$ at a 3D sample point $\mathbf{p}^d$ in deformed space using inverse distance weighting (IDW):
\begin{equation}
    \mathbf{p}^c = \mathbf{p}^d + \sum_{s \in \mathcal{N}} \frac{w_{s}}{\sum_{s'\in\mathcal{N}} w_{s'} } \mathbf{d}^b_{s},
    \label{eq:idw}
\end{equation}
where $\mathcal{N}$ are the $K=5$ nearest neighbors of $\mathbf{p}^d$ in $S_d$, and $w_{s} = w'_{s} - \min_{s'\in \mathcal{N}} w'_{s'} $ with $w'_{s}= \lVert \mathbf{p}^d - \mathbf{y}^d_{s} \rVert^{-1}$. 
We can then sample the canonical model at $\mathbf{p}^c$ as before: $(o,\mathbf{c}) = \mathbf{v}(\mathbf{p}^c)$. 
To remove spurious geometry that $o$ might show, we set $o(\mathbf{x})=0$ for $\mathbf{x}$ that are further than some threshold from $S_d$. 
Thus, we can now bend straight rays into the canonical model and render the interactively deformed state of the object in a realistic fashion. 

When needed, we can upsample the point cloud from the simulation to make it denser.
Unlike for rendering, we need to consider forward warping for this case. 

\section{Results}

\subsection{Dataset}
\label{sec:dataset}
\begin{figure*}
    \centering
    \resizebox{0.95\textwidth}{!}{
        \begin{tabular}{c c c c c c c}
            \includegraphics[height=2cm]{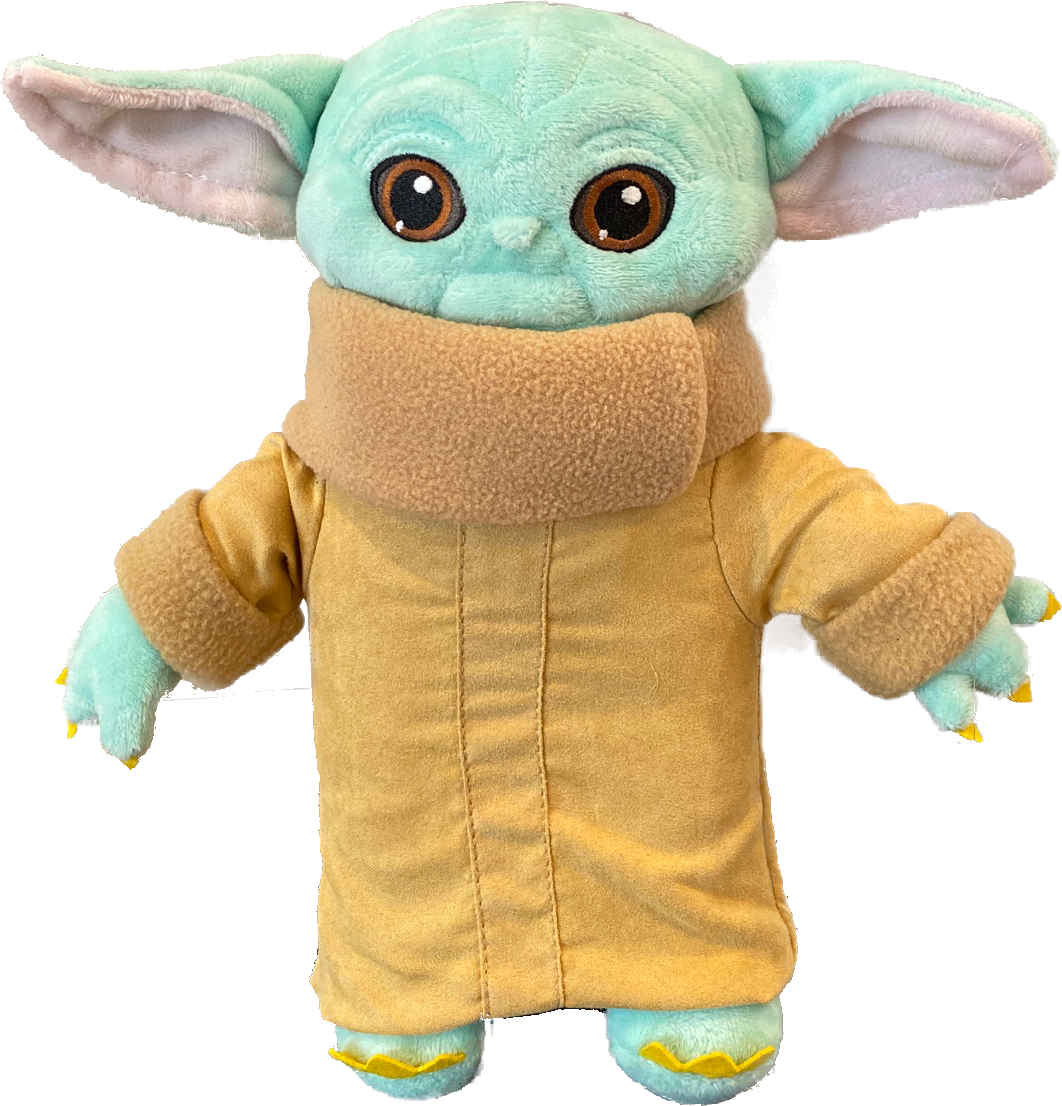} & \includegraphics[height=2cm]{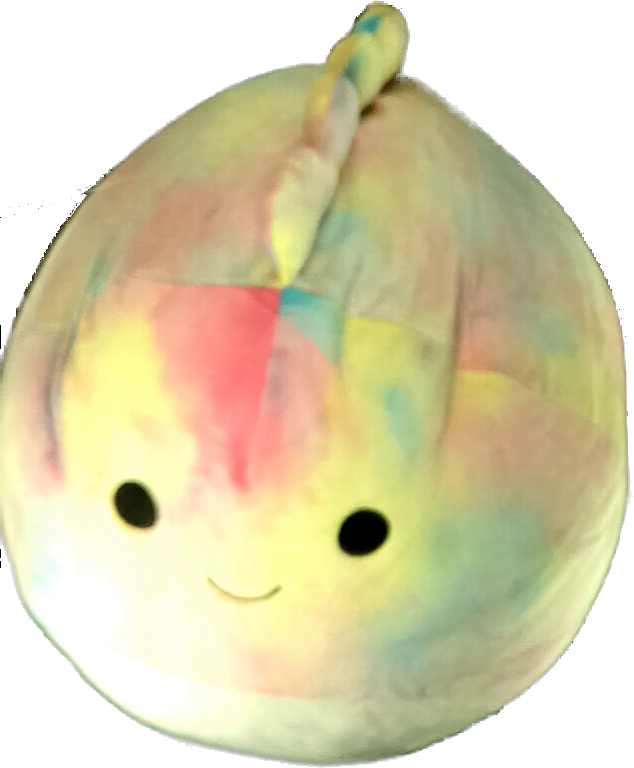} & \includegraphics[height=2cm]{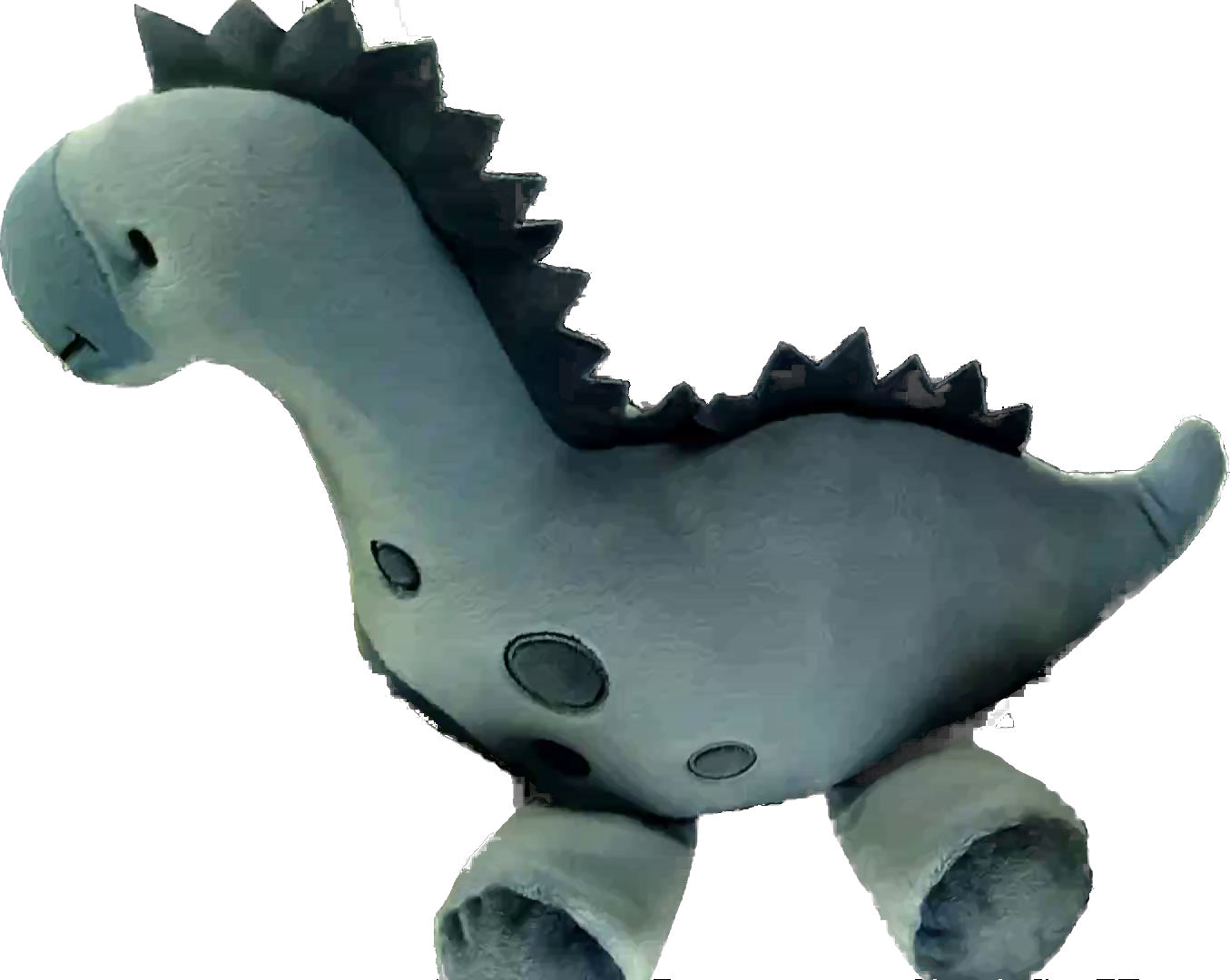} & \includegraphics[height=2cm]{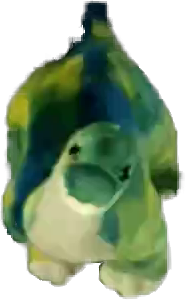} &
            \includegraphics[height=2cm]{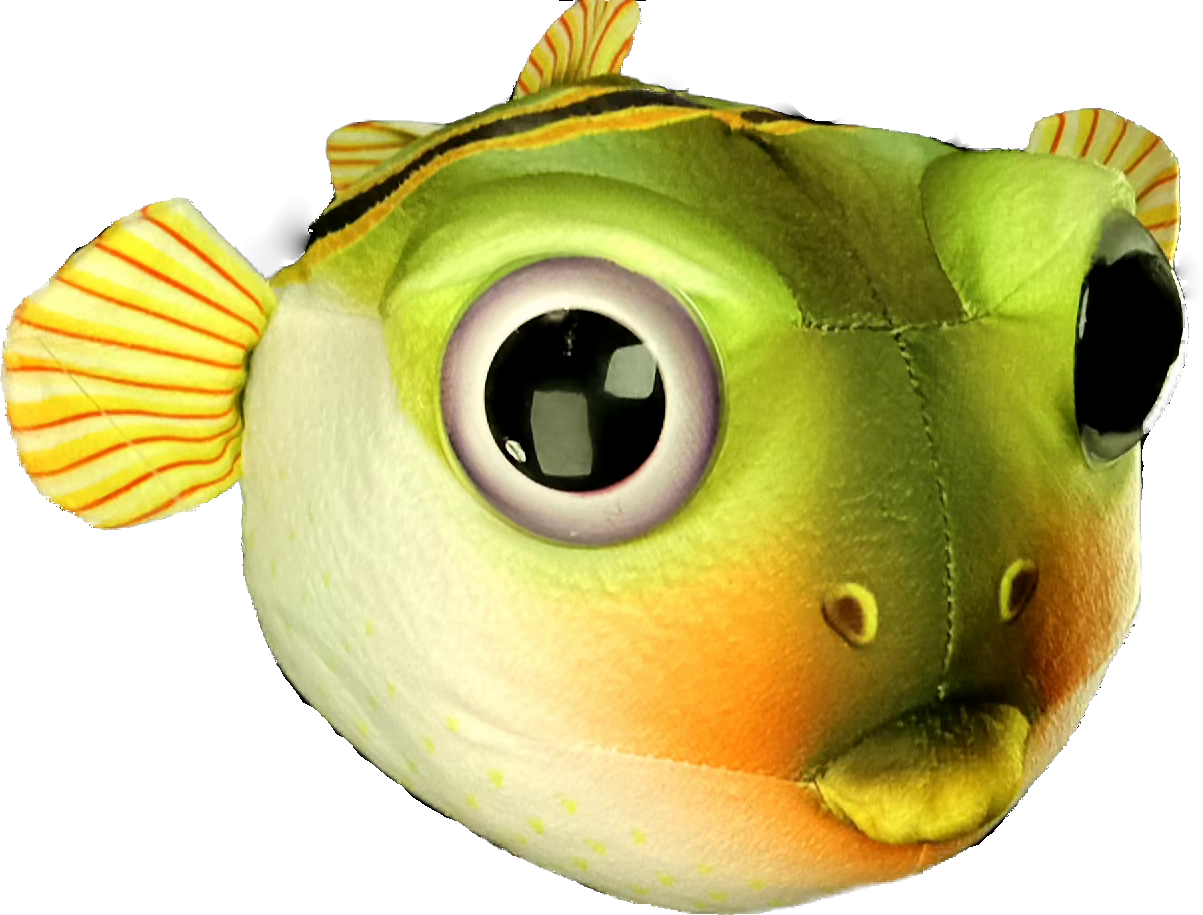} &
            \includegraphics[height=2cm]{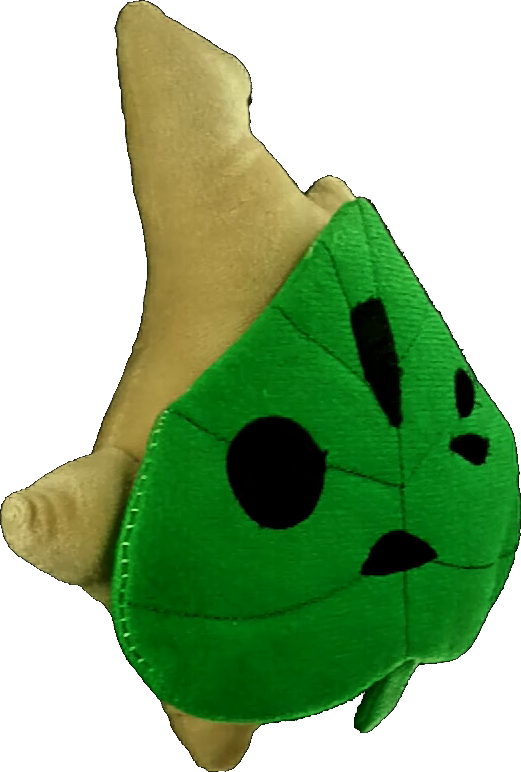} &
            \includegraphics[height=2cm]{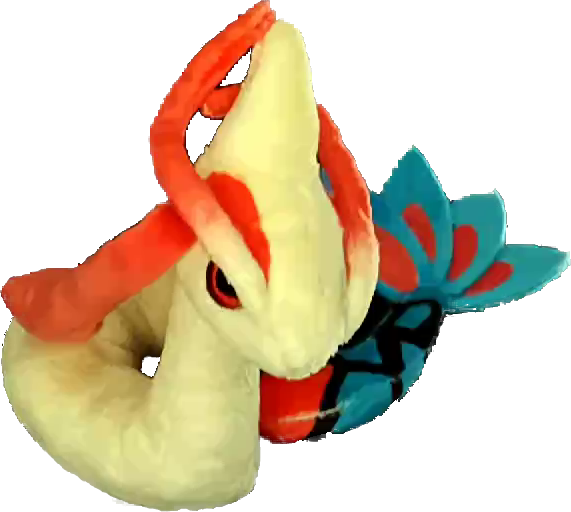} \\
            \footnotesize Baby Alien (179g, 41s)$^\|$ & \footnotesize Dino Rainbow (672g, 37s)$^\|$ & \footnotesize Dino Blue (148g, 55s)$^\|$ & \footnotesize Dino Green (76g, 42s) & \footnotesize Fish (282g, 65s)$^\|$ & \footnotesize Leaf (58g, 32s) & \footnotesize Serpentine (54g, 40s)$^*$\\
            \cline{6-7}
            \includegraphics[height=2cm]{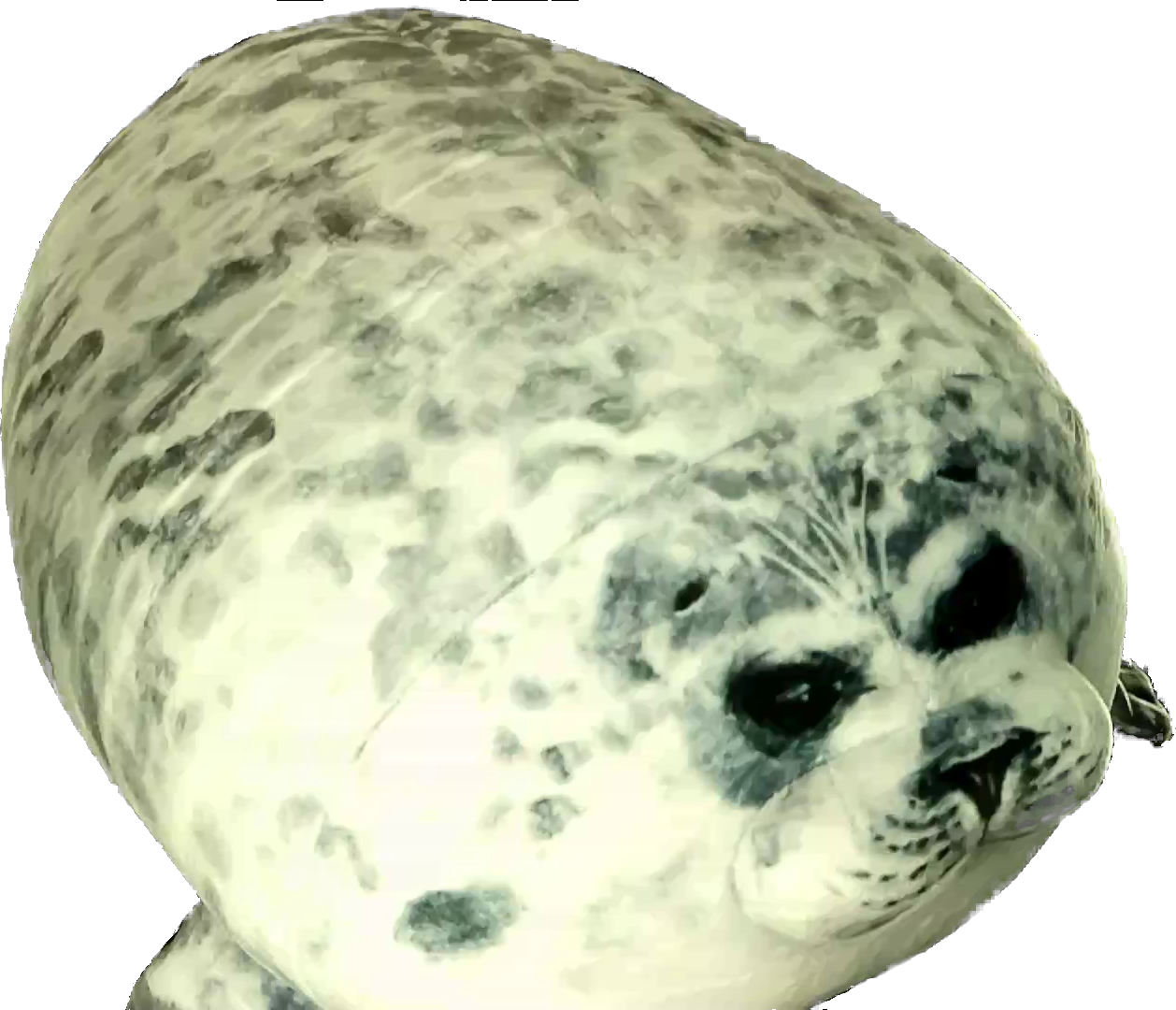} &
            \includegraphics[height=1.2cm]{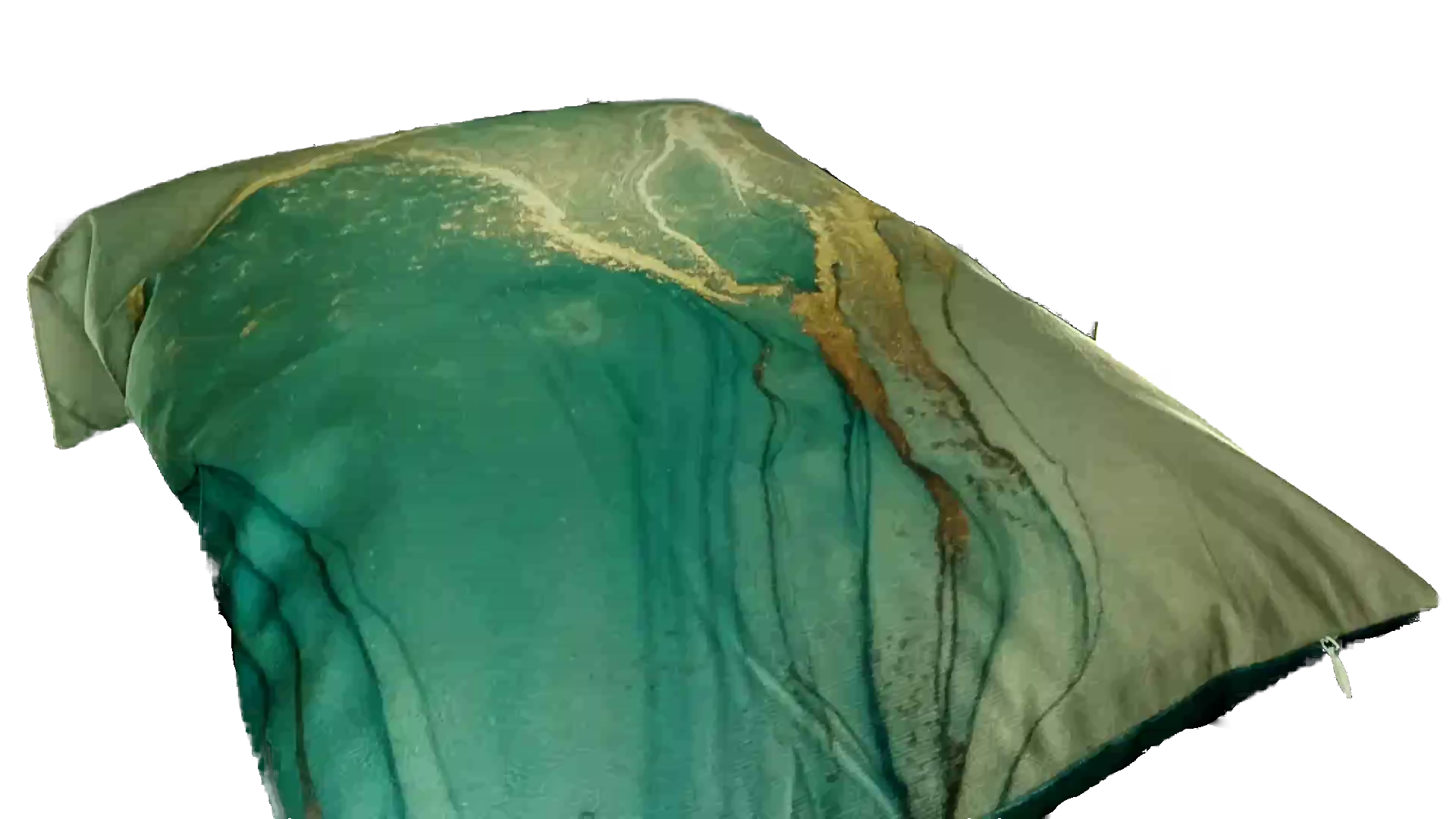} &
            \includegraphics[height=2cm]{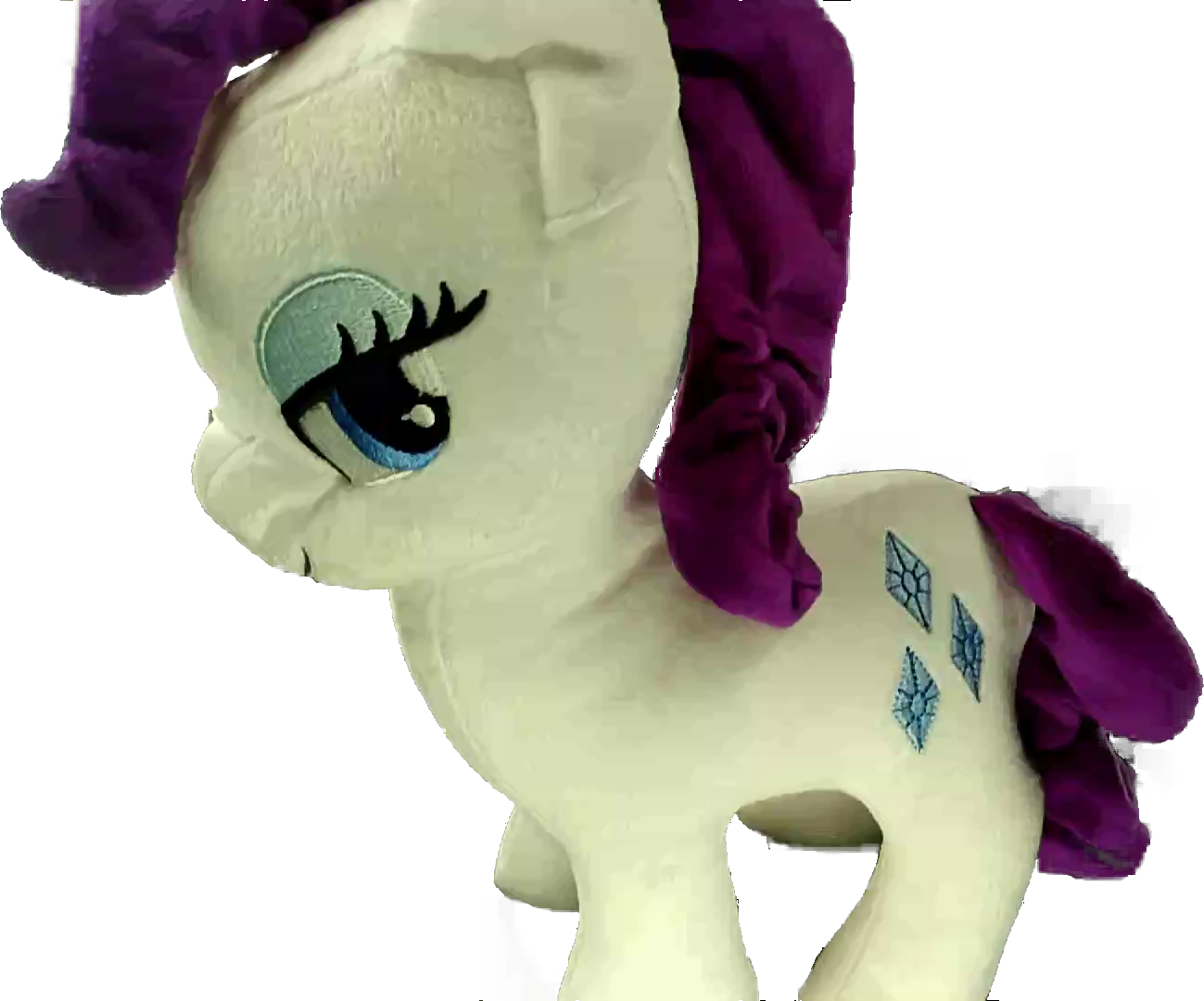} &
            \includegraphics[height=2cm]{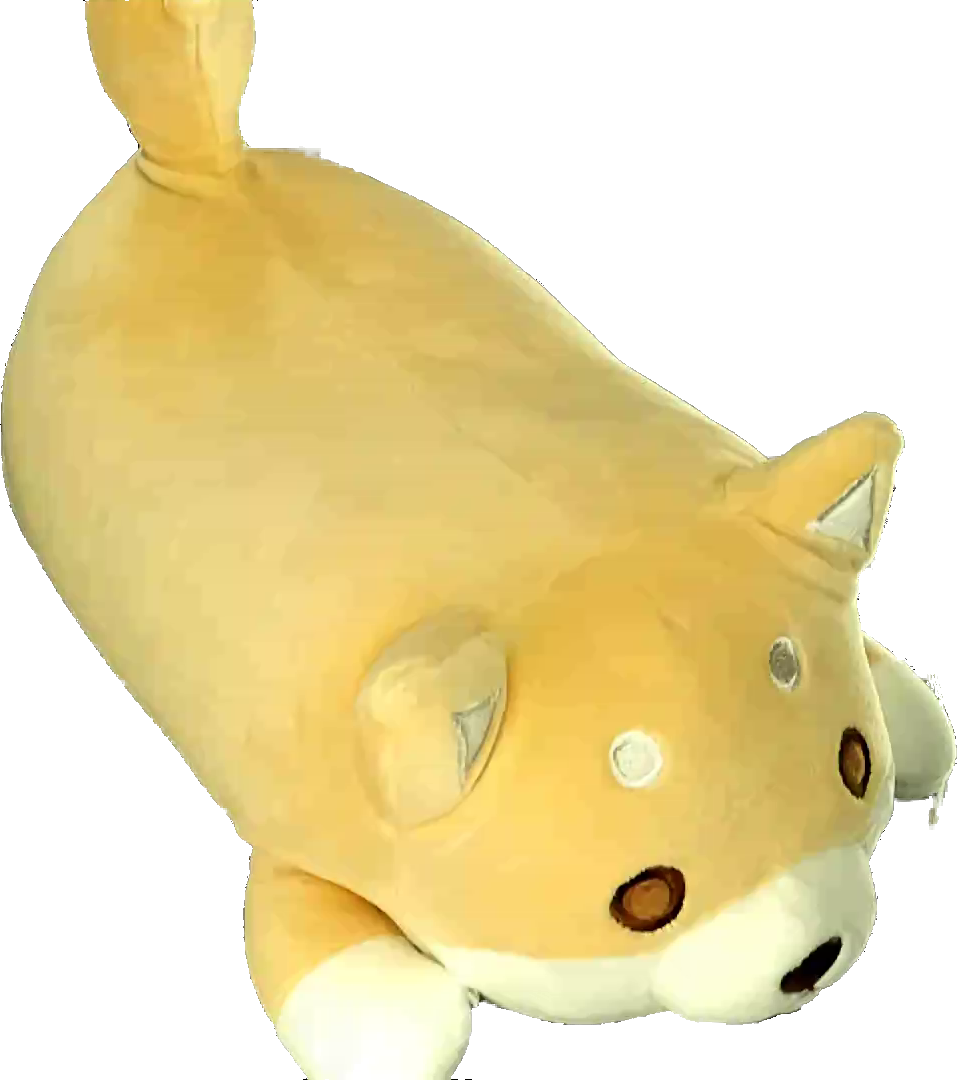} &
            \includegraphics[height=2cm]{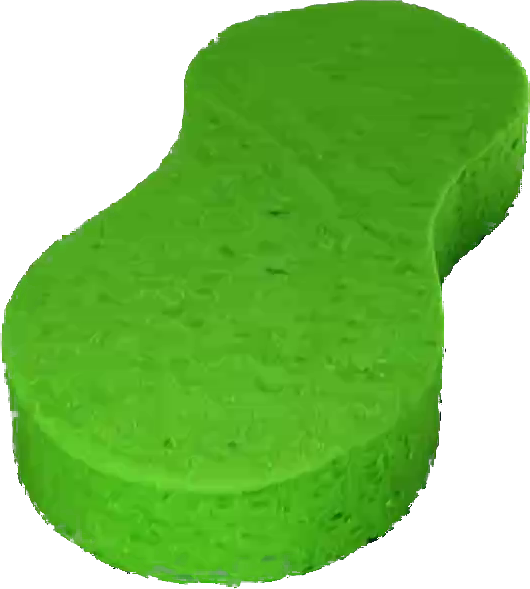} &
            \multicolumn{1}{|l}{\includegraphics[height=2.5cm,trim={10cm 5cm 10cm 0cm},clip]{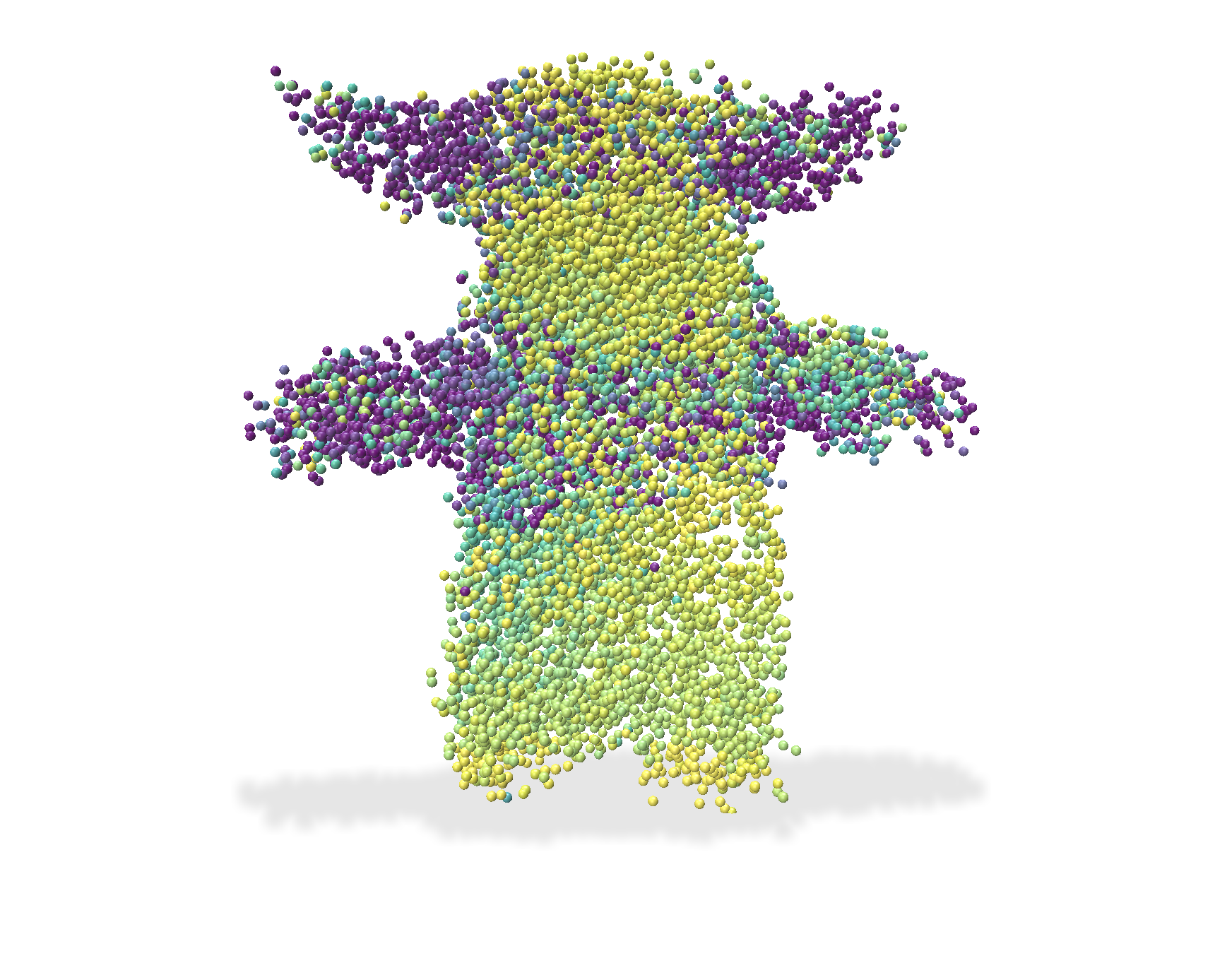}} &
            \includegraphics[height=2.5cm,trim={10cm 5cm 10cm 0cm},clip]{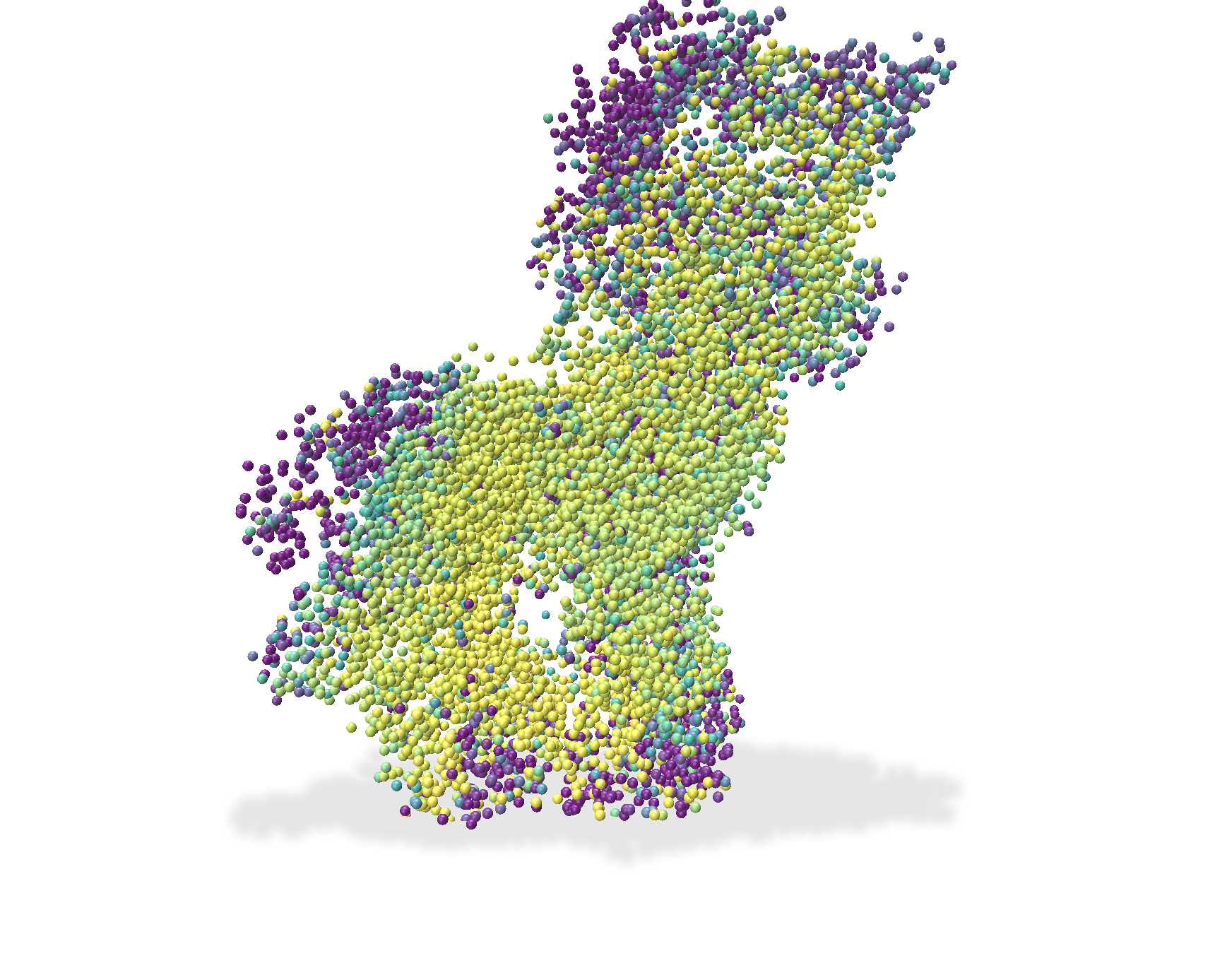} \\
            \footnotesize Mr. Seal (444g, 53s)$^\|$ &
            \footnotesize Pillow (406g, 42s) &
            \footnotesize Pony (197g, 51s)$^*$ &
            \footnotesize Dog (213g, 67s)$^\|$ &
            \footnotesize Sponge (21g, 46s) &
            \footnotesize Baby Alien \lame $\mu$ &
            \footnotesize Pony \lame  $\mu$
        \end{tabular}
    }
    \caption{\textit{The \textbf{PLUSH} dataset} consists of 12 items from everyday life: a pillow, a sponge and several plushies. $\|$ indicates that we recorded extremity motion for the object, * indicates that the recording has significant second order motion. We additionally provide the mass
    and recording duration for each object. \textbf{Lower right:} \textit{Lam\'e parameter visualizations for Baby Alien and Pony.} Colors tending towards purple show a softer region, colors tending towards green and yellow a harder region. Our method clearly identifies different material properties on the objects, for example the arms and ears for the Baby Alien, and the mane and tail of the Pony.}
    \label{fig:dataset}
\end{figure*}

The \textbf{PLUSH} dataset consists of 12 soft items encountered in everyday life (see Fig.~\ref{fig:dataset}): a pillow, a sponge, and various plush toys.
We chose items that are composed of soft (and in some cases, heterogeneous) material, complex geometry, and rich texture and color to enable successful background subtraction, 4D reconstruction and tracking. 
Our strategy for applying external forces is based on the observation that our chosen objects consist of \emph{bulk volumes} (such as the body of a plush toy) along with \emph{flexible extremities} (ears and fingers of the toy). We move object extremities by using transparent fishing line, and we use a stream of compressed air to exert force on bulk volumes. 
The nozzle position and stream direction must be tracked during video capture to provide the direction and magnitude of forces acting on the object at every point in time. 
Of the 19 cameras in our capture rig, we use three to track the nozzle using an attached ArUco marker~\cite{Garrido-Jurado2016,Romero-Ramirez2018}. 
Using this system, we generate multi-part video sequences for each capture subject, where we sequentially actuate the fishing lines (when applicable) followed by sweeping the air stream over the object.
We record between 32s and 67s of video for each object, at a frame rate of 40FPS.

\subsection{Virtual Elastic Objects}

\begin{table}
    \centering
    \resizebox{0.4\textwidth}{!}{
    \begin{tabular}{c|c|c|c}
        Object & average (mm) & 95\% (mm) & max (mm) \\
        \hline
        Baby Alien   & 3.8 & 14.4 & 29.3 \\
        \rowcolor{Gray}
        Fish        & 1.1 & 6.6 & 18.5 \\
        Leaf       & 0.4 & 1.1 & 9.8 \\
        \rowcolor{Gray}
        Mr. Seal    & 0.4 & 1.9 & 171.9 \\
        Pillow & 1.5 & 7.8 & 18.35 \\
        \rowcolor{Gray}
        Dog       & 1.7 & 7.5 & 28.8 \\
        Sponge      & 0.2 & 1.8 & 15.8 \\
        \rowcolor{Gray}
        Dino Rainbow     & 4.0 & 14.6 & 171.4 \\
        Dino Blue   & 5.5 & 56.0 & 105.8\\
        \rowcolor{Gray}
        Dino Green  & 6.2 & 68.4 & 132.0 \\
        Pony        & 21.1 & 164.3 & 204.9 \\
        \rowcolor{Gray}
        Serpentine     & 7.5 & 43.1 & 94.7 \\
        \hline
        \hline
        Average*    & 2.5 & 18.0 & 70.2 \\
        Average     & 4.4 & 32.3 & 83.5 \\
        \hline
    \end{tabular}
    }
    \caption{\textit{$\ell_2$ distance of simulated point clouds compared with reconstructed point clouds on the test set.} We record the average distance per point per frame, the 95th percentile of average point distances of all frames, and the maximum distance of all points. Average* excludes the data from Pony and Serpentine.}
    \label{tab:distance}
\end{table}

For each of the 12 examples, we create a VEO using 20-50 frames from the reconstruction and evaluate on the remaining 500-1500 frames.
We use the $\ell_2$ distance between the surface points of the VEO to the reconstructed point cloud from the captured data to evaluate the quality of the reconstructed parameters.
For all examples except for the Baby Alien we use the external force field data obtained using the air stream.
For the Baby Alien, we specifically use the arm and ear motion to demonstrate the versatility of our method in this scenario.
We present the results in Tab.~\ref{tab:distance}.

The error is relatively small for all objects, which shows that our method is applicable to objects with different geometries, and can learn the corresponding material parameters even for heterogeneous objects.
Larger errors are observed for objects with a thin and tall component (see the last 4 rows of the table).
This error is largely caused by tracking inaccuracies of the nozzle: even slight inaccuracies can cause large errors when, for example, the neck of the dinosaur moves while the recorded air stream direction does not, or barely, touch the object.

\begin{figure}
    \centering
    \includegraphics[width=0.4\textwidth]{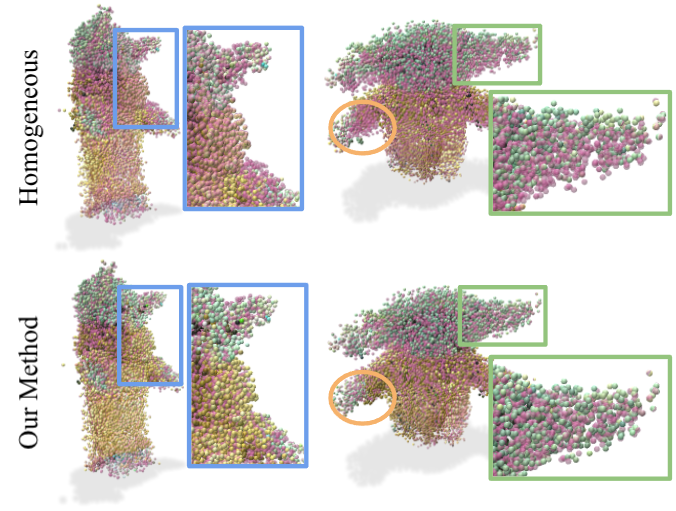}
    \caption{\textit{Baseline comparison with a homogeneous material.} We show two postures for both material settings, overlayed over the ground truth point cloud in purple. The homogeneous parameters have been optimized. The inhomogeneous model has clear advantages over the homogeneous model: the core body posture is better simulated, and the arms and floppy ears are better posed.}
    \label{fig:errorMap}
    \vspace*{0.2cm}
\end{figure}

\noindent\textbf{Inhomogeneous Material.}
An important feature of our method is that it can identify different material parameters for different parts of the object (c.t. Fig.~\ref{fig:dataset}, lower right).
This is crucial for building a detailed physics model with no prior knowledge of the object.
Even more, our method can reliably learn `effective' softness of the material even in places with unreliable tracking, for example thin geometrical structures close to joints.
In case of Baby Alien, our method learns that the ears and arms are softer compared to the other body parts; the mane and tail of the Pony are softer, even though these regions are very hard to track.
Both reconstructions match the properties of their real counterparts.

We show a comparison between our method that assumes an independent material parameter on all points with a baseline with only one global material parameter.
We trained the baseline model with the exact same procedure as before but learn just one $\mu$ and one $\lambda$ for the energy in Eq.~\ref{eq:neoHookean}.
As shown in Fig.~\ref{fig:errorMap}, our inhomogeneous model is visually indistinguishable from the ground truth point cloud, while the homogeneous baseline model has a larger error.
The homogeneous model fails to capture the exact movements at the arms and ears of the Baby Alien, but instead distributes the deformation evenly at the ears and arms.

\begin{figure}
    \centering
    \includegraphics[width=4cm,trim={10cm 8cm 10cm 7cm},clip]{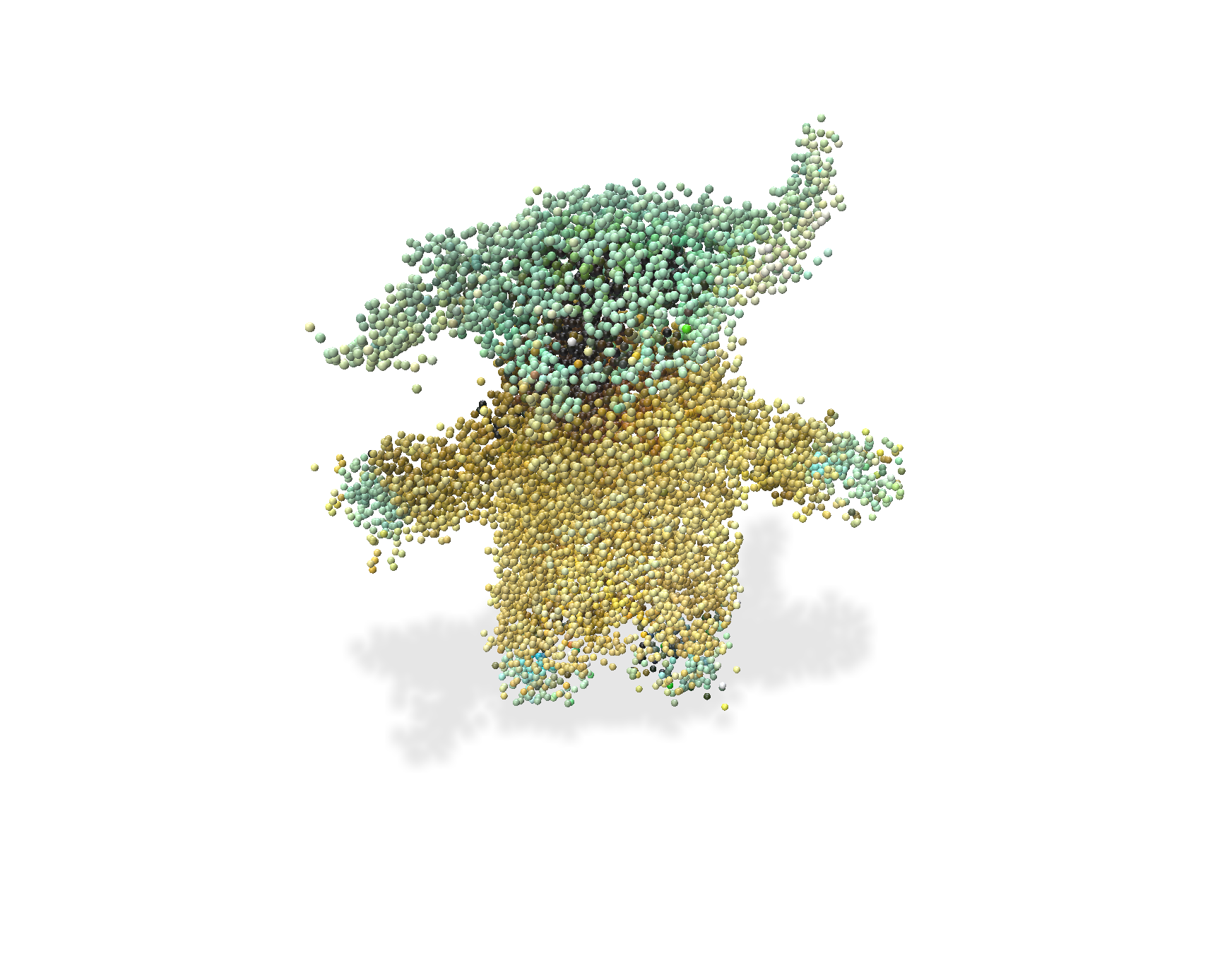}
    \includegraphics[width=4cm,trim={10cm 5cm 10cm 7cm},clip]{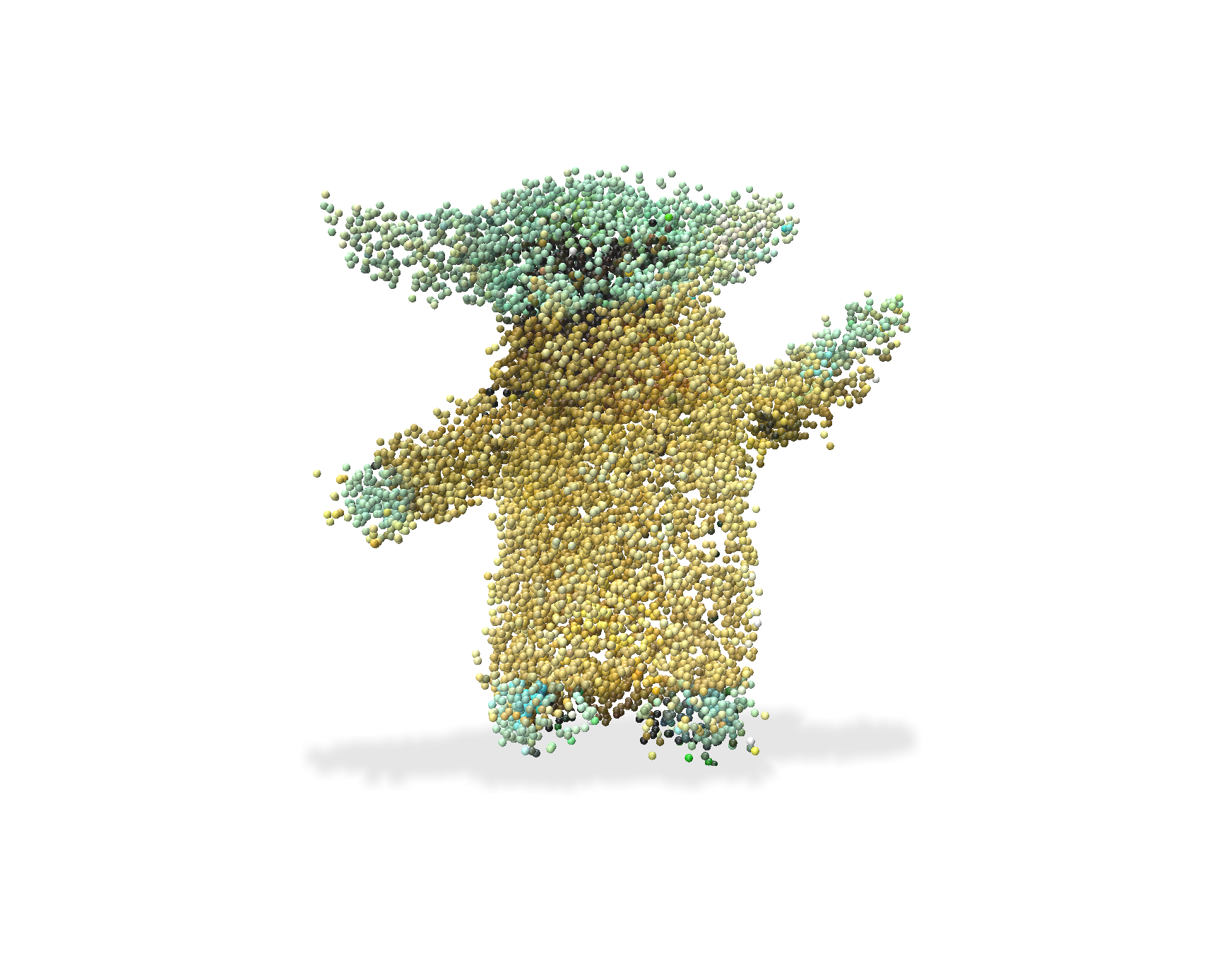}
    \vspace*{-0.2cm}
    \caption{\textit{Simulation of Baby Alien in poses unseen in the dataset.} Using the material model and simulator our method generalizes well to these asymmetric postures for ears and arms; we only observe symmetric forward and backward motions during training.}
    \label{fig:novel}
\end{figure}

\noindent\textbf{Generalization to Novel Poses.}
The strength of the underlying physics simulator is the ability to generalize to scenarios that are not encountered in the training set.
We show different simulated poses of the Baby Alien in Fig.~\ref{fig:novel}, such as pulling the ears in opposite directions, and moving just one single arm.
This deformation is particularly challenging for purely data-driven methods since both ears and arms only move synchronously in the training data.

\begin{figure}
    \centering
    \includegraphics[width=0.23\textwidth,trim={0cm 4cm 0cm 10cm},clip]{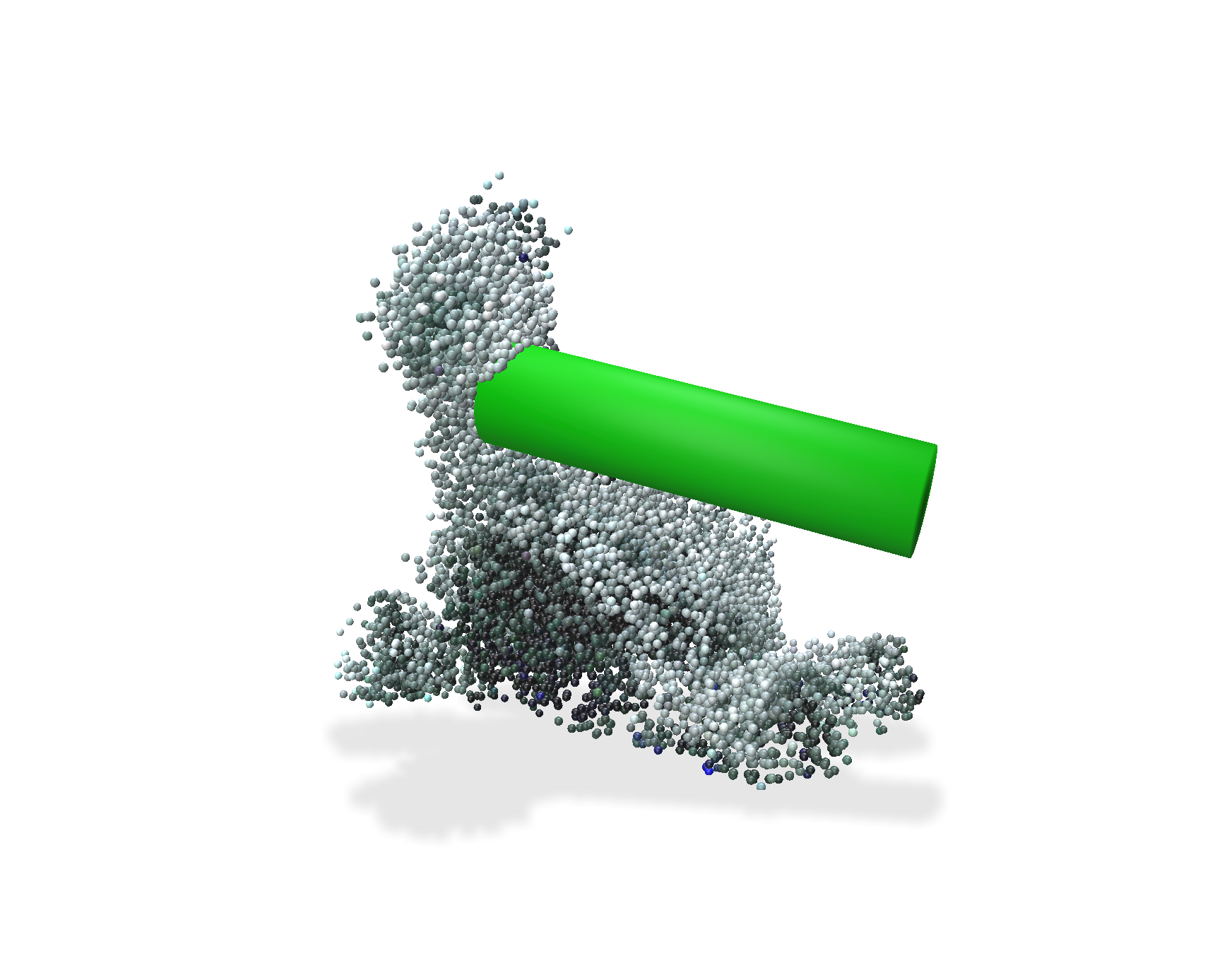}
    \includegraphics[width=0.23\textwidth,trim={0cm 4cm 0cm 10cm},clip]{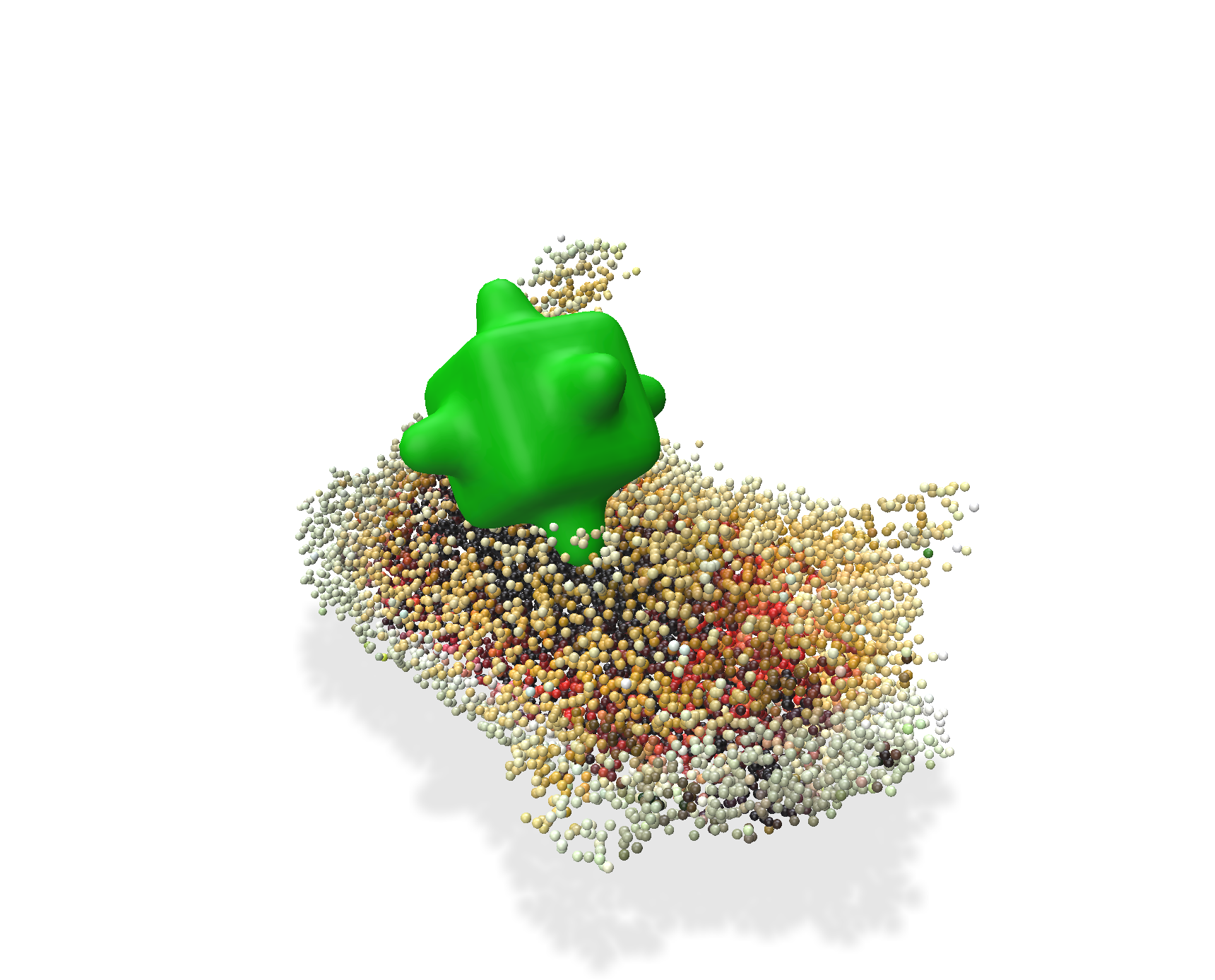}\\
    \includegraphics[width=0.23\textwidth]{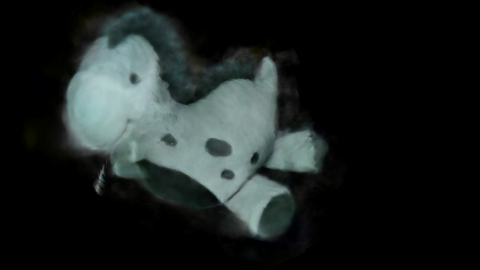}
    \includegraphics[width=0.23\textwidth]{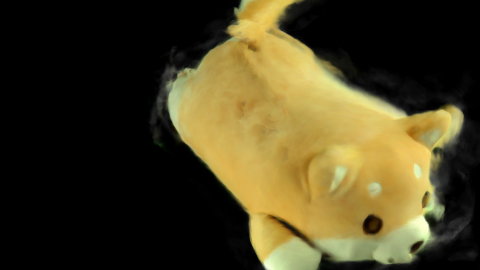}
    \caption{\textit{Rendering of the Dino Blue and Dog VEOs during interactions with secondary objects.} The dinosaur neck bends correctly, and dents are forming on the Dog's back.}
    \label{fig:collision-and-render}
\end{figure}
\noindent\textbf{Interaction with Virtual Objects.}
The physical model of the object enables interactions with all kinds of different virtual items.
Fig.~\ref{fig:collision-and-render} shows the one-way coupled interaction of the learned elastic objects with other virtual items.

\noindent\textbf{Rendering}
Our pipeline ends with re-rendering an object under novel interactions not seen during training. 
Fig.~\ref{fig:collision-and-render} contains renderings of the Dino Blue and Dog objects, including interactions with two virtual objects. 
Tab.~\ref{tab:rerender} contains quantitative results, where we compare the renderings obtained from the reconstructed point clouds (which are used for supervision when learning the material parameters) and the simulated point clouds. 
The former thus provides a soft upper bound of the quality that the simulator can achieve.
We find that the simulator results are very close to those from the reconstructed point clouds. 
Thus, both the quantitative and qualitative results show that our approach is able to synthesize renderings of novel deformed states in a realistic manner. 
\iftrue
\begin{table}
    \centering
    \renewcommand{\arraystretch}{0.9} 
    \setlength{\tabcolsep}{2pt} 
    \resizebox{0.47\textwidth}{!}{
    \begin{tabular}{c|c|c|c|c|c|c|c|c|c|c|c|c|}
         & \multicolumn{6}{|c|}{Simulated} & \multicolumn{6}{|c|}{Reconstructed} \\
         & \multicolumn{3}{|c|}{Not Masked} & \multicolumn{3}{|c|}{Masked} & \multicolumn{3}{|c|}{Not Masked} & \multicolumn{3}{|c|}{Masked} \\
         Object & PSNR & SSIM & LPIPS & PSNR & SSIM & LPIPS & PSNR & SSIM & LPIPS & PSNR & SSIM & LPIPS \\
        \hline
        Baby Alien   & 18.40 & 0.734 & 0.255 & 21.17 & 0.840 & 0.174 & 18.75 & 0.747 & 0.249 & 21.92 & 0.853 & 0.167 \\
        \rowcolor{Gray}
        Fish        & 19.75 & 0.692 & 0.239 & 22.55 & 0.808 & 0.173 & 20.03 & 0.701 & 0.235 & 22.96 & 0.818 & 0.169 \\
        Leaf       & 25.14 & 0.901 & 0.091 & 27.32 & 0.935 & 0.065 & 25.19 & 0.901 & 0.091 & 27.37 & 0.935 & 0.065 \\
        \rowcolor{Gray}
        Mr. Seal    & 20.61 & 0.697 & 0.240 & 24.03 & 0.801 & 0.180 & 20.65 & 0.698 & 0.239 & 24.11 & 0.802 & 0.180 \\
        Pillow      & 21.45 & 0.743 & 0.223 & 23.18 & 0.806 & 0.174 & 21.92 & 0.760 & 0.218 & 23.84 & 0.823 & 0.169 \\
        \rowcolor{Gray}
        Dog       & 18.98 & 0.751 & 0.206 & 24.68 & 0.904 & 0.104 & 19.05 & 0.757 & 0.203 & 25.24 & 0.912 & 0.100 \\
        Sponge      & 21.94 & 0.846 & 0.130 & 26.99 & 0.925 & 0.070 & 21.92 & 0.846 & 0.130 & 27.01 & 0.925 & 0.070 \\
        \rowcolor{Gray}
        Dino Rainbow      & 18.64 & 0.754 & 0.302 & 23.87 & 0.839 & 0.232 & 20.22 & 0.778 & 0.281 & 26.21 & 0.859 & 0.213 \\
        Dino Blue   & 18.48 & 0.702 & 0.244 & 20.70 & 0.848 & 0.160 & 19.56 & 0.726 & 0.227 & 22.06 & 0.871 & 0.143 \\
        \rowcolor{Gray}
        Dino Green  & 18.94 & 0.779 & 0.190 & 21.49 & 0.863 & 0.135 & 20.46 & 0.794 & 0.180 & 23.59 & 0.879 & 0.121 \\
        Pony        & 16.54 & 0.758 & 0.245 & 19.20 & 0.859 & 0.163 & 19.31 & 0.798 & 0.200 & 24.65 & 0.906 & 0.108 \\
        \rowcolor{Gray}
        Serpentine     & 18.22 & 0.798 & 0.181 & 21.39 & 0.903 & 0.111 & 19.95 & 0.813 & 0.162 & 23.14 & 0.916 & 0.091 \\
        \hline
        \hline
        Average*    & 20.23 & 0.760 & 0.212 & 23.60 & 0.857 & 0.145 & 20.78 & 0.771 & 0.205 & 24.43 & 0.868 & 0.140 \\
        Average     & 19.76 & 0.763 & 0.212 & 23.05 & 0.861 & 0.147 & 20.58 & 0.777 & 0.201 & 24.34 & 0.875 & 0.133 \\
    \end{tabular}
    }
    \caption{\emph{Rendering evaluation.} We report the classic error metrics PSNR and SSIM~\cite{wang2004image} ($-1$ to $+1$), where higher is better for both, and the learned perceptual metric LPIPS~\cite{zhang2018unreasonable} (0 is best). We use deformed point clouds to render deformed states of the canonical model, see Sec.~\ref{ssec:rendering}. We use both the point cloud $P_t$ that the reconstruction (Sec.~\ref{ssec:recon}) provides directly (`Reconstructed') or the point cloud that the simulator provides after learning the material parameters (Sec.~\ref{ssec:learning_material}, `Simulated'). We report two versions: we either apply the segmentation masks of the input images to the rendered image to remove all artifacts that spill-over onto the background (`Masked') or we do not (`Not Masked'). Note that the values on the reconstructed point cloud are a (soft) upper bound for what the simulator can achieve. The simulated results are close the reconstructed results, demonstrating that the learned material parameters yield deformation fields that allow to re-render the object as well as the reconstruction can.
    }
    \label{tab:rerender}
\end{table}
\fi

\section{Limitations} 
\noindent\textbf{Artifacts.} 
Due to the sparse camera setup (16 cameras for 360 degree coverage), we found NeRF unable to reconstruct viewpoint dependent effects, leading to artifacts around specular regions like eyes. 
Furthermore, the air compressor leads to quickly oscillating surfaces (\eg, the fins of the fish), which pose a challenge for reconstruction and material parameter estimation, and impacts calibration. 
These issues impact the extracted point clouds as well as the final renderings (artifacts visible in Fig.~\ref{fig:collision-and-render}); we manually removed resulting background clutter in the point clouds.
The physical simulator turned out to be remarkably robust towards noise and can run with any reconstructed point cloud with temporal correspondences.

\noindent\textbf{Known Forces.} 
The simulator requires the forces impacting the object during capture to be known. 
This limits the variety of forces that can be applied and hence the kind of objects that are compatible with the presented method. 
We expect an extension to handle unknown forces to be a challenging but exciting direction for future work.
Finding good force priors could be a viable approach in this direction.

\section{Conclusion} 
We introduced a novel, holistic problem setting: estimating physical parameters of a general deformable object from RGB input and known physical forces, and rendering its physically plausible response to novel interactions realistically. 
We further proposed Virtual Elastic Objects as a solution and demonstrated their ability to synthesize deformed states that greatly differ from observed deformations. 
Our method leverages a physical simulator that is able to estimate plausible physical parameters from a 4D reconstruction of a captured object. 
Finally, we showed that these deformed states can be re-rendered with high quality. 
We hope that the presented results and the accompanying dataset will inspire and enable future work on reconstructing and re-rendering \emph{interactive} objects.
\newpage

{\small
\bibliographystyle{ieee_fullname}
\bibliography{reference}
}

\end{document}